\documentclass[10pt,twocolumn,letterpaper]{article}

\usepackage{cvpr}              
\usepackage[accsupp]{axessibility} 

\usepackage{stfloats}
\usepackage[dvipsnames]{xcolor}

\newcommand{\supp}{\textit{Supplementary Material}\xspace}

\definecolor{cvprblue}{rgb}{0.21,0.49,0.74}
\usepackage[pagebackref,breaklinks,colorlinks,citecolor=cvprblue]{hyperref}

\def\paperID{1692} 
\def\confName{CVPR}
\def\confYear{2024}

\usepackage{colortbl}
\usepackage{booktabs,multirow}
\usepackage{makecell}
\usepackage{tabulary}
\usepackage{fontawesome5}
\usepackage{bbding}
\newcommand \blfootnote[1]{
    \begingroup
        \renewcommand
        \thefootnote{}\footnote{#1}
        \addtocounter{footnote}{-1}
        \vspace{-1ex}
    \endgroup
}
\newlength\savewidth\newcommand\shline{\noalign{\global\savewidth\arrayrulewidth
  \global\arrayrulewidth 1pt}\hline\noalign{\global\arrayrulewidth\savewidth}}

\newcommand{\tablestyle}[2]{\setlength{\tabcolsep}{#1}\renewcommand{\arraystretch}{#2}\centering\footnotesize}
\newcommand{\method}{\texttt{SimM}\xspace}
\newcommand{\benchmark}{SimMBench\xspace}

\title{Check, Locate, Rectify: A Training-Free Layout Calibration System for Text-to-Image Generation}

\author{Biao Gong$^{1}$\footnotemark[2]\>\,$^{\text{\Envelope}}$, Siteng Huang$^{2}$\footnotemark[2]\>\,\textsuperscript{*}, Yutong Feng$^{1}$, Shiwei Zhang$^{1}$, Yuyuan Li$^{2}$, Yu Liu$^{1}$\\
{$^1$Alibaba Group\ \ $^2$Zhejiang University}\\[-2px]
{\tt\small \{a.biao.gong, siteng.huang\}@gmail.com y2li@zju.edu.cn}\\[-3px]
{\tt\small \{fengyutong.fyt, zhangjin.zsw, ly103369\}@alibaba-inc.com}
}

\begin{document}
\twocolumn[{
\maketitle
\vspace{-8mm}
\begin{center}
    \captionsetup{type=figure}
        \vspace{-1mm}
    \includegraphics[width=\linewidth]{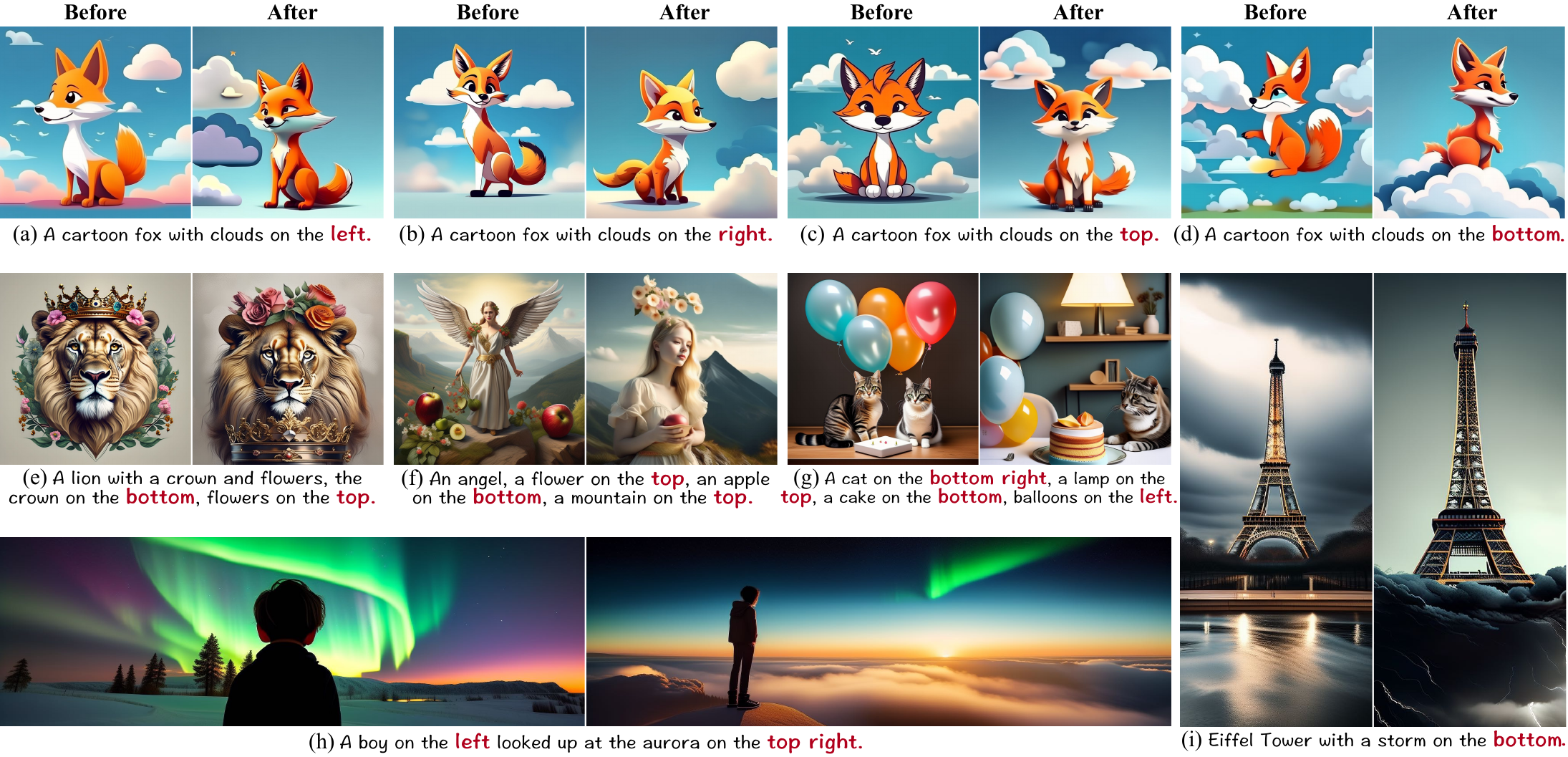}
    \vspace{-6mm}
    \captionof{figure}{
    Given only the input textual prompt, our system can autonomously detect and rectify the layout inconsistencies across various \textbf{position requirements} (a-d), \textbf{object quantities} (e-g), and \textbf{resolutions} (h-i).
    }
    \label{fig:teaser}
\end{center}
}]
\blfootnote{$^\dagger$Equal contribution. $^{\text{\Envelope}}$ Corresponding author.}
\blfootnote{\textsuperscript{*}Work done during internship at Alibaba Group.}

\begin{abstract}
Diffusion models have recently achieved remarkable progress in generating realistic images.
However, challenges remain in accurately understanding and synthesizing the layout requirements in the textual prompts.
To align the generated image with layout instructions, we present a training-free layout calibration system \method that intervenes in the generative process on the fly during inference time.
Specifically, following a ``check-locate-rectify'' pipeline, the system first analyses the prompt to generate the target layout and compares it with the intermediate outputs to automatically detect errors.
Then, by moving the located activations and making intra- and inter-map adjustments,
the rectification process can be performed with negligible computational overhead.
To evaluate \method over a range of layout requirements, we present a benchmark \benchmark that compensates for the lack of superlative spatial relations in existing datasets.
And both quantitative and qualitative results demonstrate the effectiveness of the proposed \method in calibrating the layout inconsistencies. Our project page is at \url{https://simm-t2i.github.io/SimM}.
\end{abstract}

\section{Introduction}

\begin{figure}[!t]    
  \centering
   \includegraphics[width=0.95\linewidth]{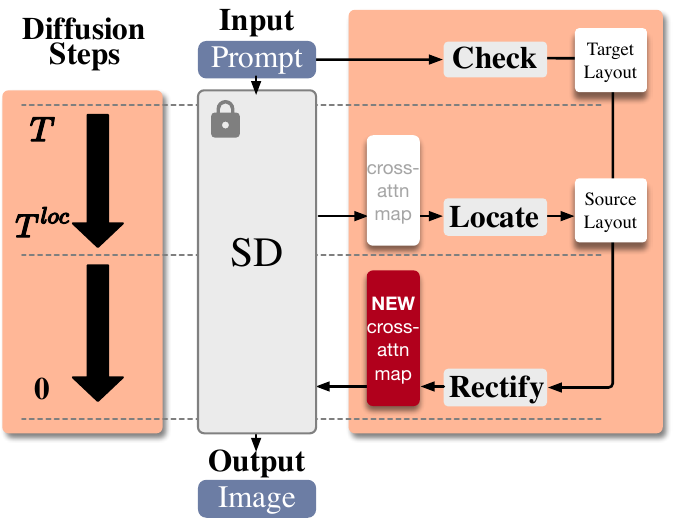}
   \vspace{-2mm}
   \caption{
   \textbf{The ``check-locate-rectify'' pipeline of \method}, intervening in the generative process on the fly during inference.
   }
   \vspace{-5mm}
   \label{fig:pipeline}
\end{figure}

Text-to-image generation~\cite{Reed:GAN-T2I,Ramesh:DALL-E,Li:GLIGEN,Ge:region-based-diffusion} has emerged as a promising application of AI-generated content (AIGC), demonstrating the remarkable ability to generate synthetic images from conditional text descriptions.
This technology has attracted considerable attention in recent years due to its potential impact on various domains such as image customization~\cite{Ruiz:DreamBooth,Zhang:ControlNet}, 3D content creation~\cite{Poole:DreamFusion,Lin:Magic3D} and virtual reality~\cite{Bussell:T2I-VR}.
Since achieving high-quality and diverse image generation is challenging,
recent advancements have witnessed the rise of diffusion models~\cite{Ho:diffusion,Rombach:Stable-Diffusion}.
Diffusion models employ a sequential generation process that gradually refines the generated images by iteratively conditioning on noise variables.
This iterative refinement mechanism allows for an improvement in the fidelity and quality.

Despite the effectiveness of diffusion models, a significant challenge remains: 
most text-to-image generators, typified by Stable Diffusion~\cite{Rombach:Stable-Diffusion}, show limitations in accurately understanding and interpreting textual layout instructions~\cite{Gokhale:T2I-spatial-benchmark}.
This can be regarded as a kind of ``hallucination''~\cite{Gunjal:VLM-hallucination,Zhou:VLM-hallucination}, which refers to the phenomenon that the generated image is inconsistent with the prompt content.
On the one hand, various textual descriptions include the relative relation ``\textit{a dog to the left of a cat}'' and the superlative relation ``\textit{the crown on the bottom}'', presenting an inherent difficulty for automated systems to parse and understand layout information.
Besides, inaccuracies in spatial relations may be due to the prior knowledge embedded in pre-trained models, as the large dataset may contains certain biases or assumptions about object placement or orientation.
To exemplify this point, consider the following situation: since the ``\textit{crown}'' in the training images are predominantly positioned over the head of another organism, it becomes difficult to specify their occurrence below~(\cref{fig:teaser}-e).

These factors not only compromise the quality and fidelity of the generated images but also hinder the overall utility and user experience of text-to-image generation systems.
Some efforts~\cite{Yang:ReCo,Zheng:LayoutDiffusion} attempt to address the issue by training auxiliary modules or fine-tuning diffusion models on datasets with layout annotations.
Apart from the difficulty of collecting sufficient high-quality data, these resource-intensive methods require retraining for each given checkpoint, 
making them struggle to keep up with the rapid version iterations of base models.

In this paper, we delve into the exploration of layout calibration given a pre-trained text-to-image diffusion model.
Consequently, we present a training-free real-time system \method, which follows the proposed ``\textit{check-locate-rectify}'' pipeline.
The \textbf{checking} stage is first applied to mitigate the potential impact on the generation speed, where \method generates approximate target layout for each object by parsing the prompt and applying heuristic rules.
After comparing the target layout with the intermediate cross-attention maps, layout rectification can be initiated if there are layout inconsistencies, and \method locates the misplaced objects during the \textbf{localization} stage.
Finally, during the \textbf{rectification} stage, \method transfers the located activations to the target regions, and further adjusts them with intra-/inter-map activation enhancement and suppression.
The entire workflow only affects the generation process, avoiding any additional training or loss-based updates.

We conduct both quantitative and qualitative experiments to evaluate the effectiveness of the proposed \method.
Since the popular DrawBench dataset~\cite{Saharia:DrawBench} only contains prompts with relative spatial relations, we present a new benchmark \benchmark that includes superlative descriptions composed of various orientations and objects, compensating for the diversity of textual prompts.
Compared to the recent works~\cite{Zheng:LayoutDiffusion,Chen:forward-and-backward,Phung:Attention-Refocusing}, which rely on precise target layout provided by the user, 
\method achieves satisfactory correction results even when the target layout is not precise enough, leading to a significant improvement in the layout fidelity of the generated images.

\section{Methodology}

\begin{figure*}[!t]    
  \centering
   \includegraphics[width=0.99\linewidth]{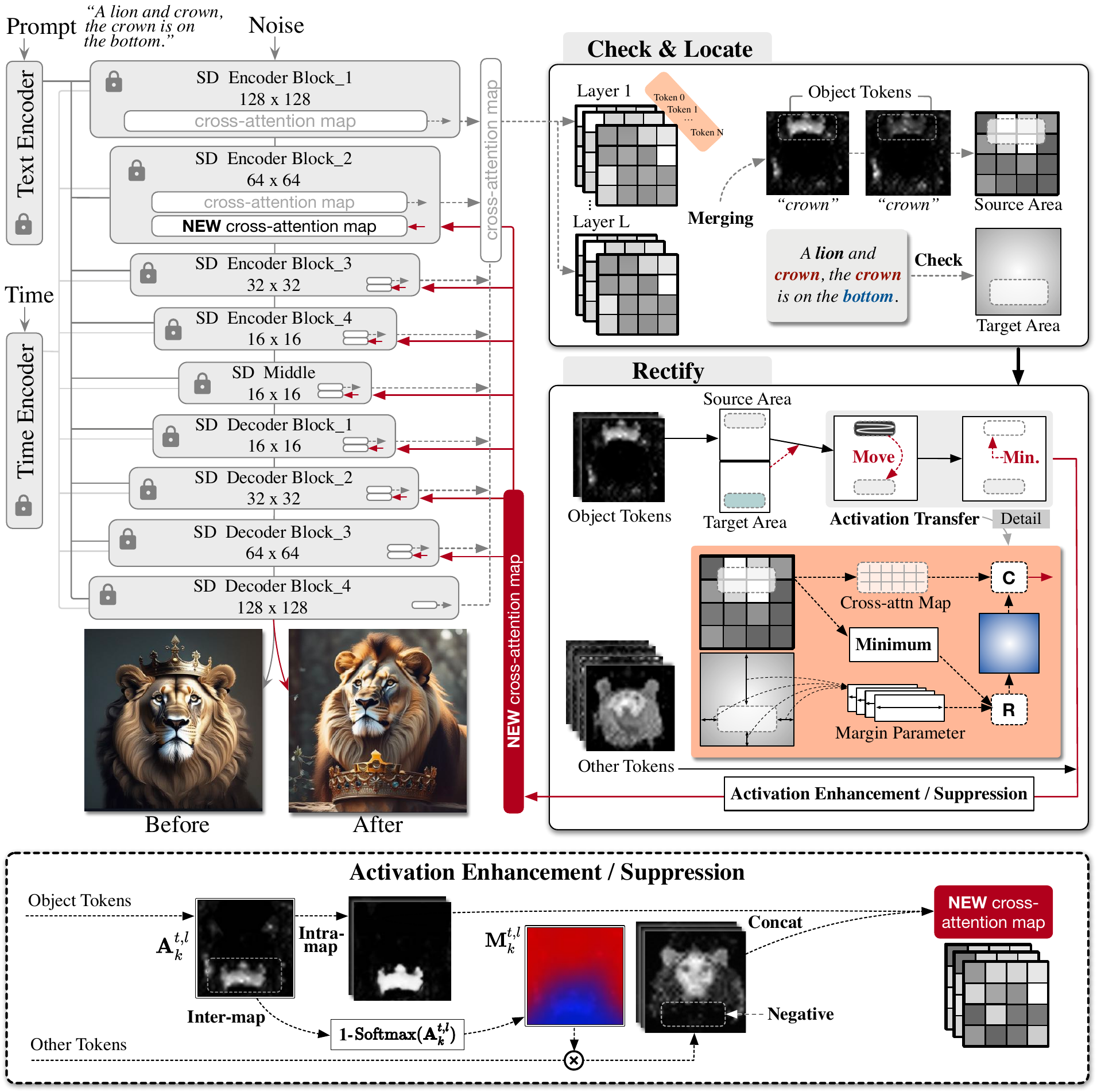}
   \vspace{-2mm}
   \caption{
   \textbf{A detailed illustration of our \method system.} \textit{R} means repeating, \textit{C} means concatenating.
   }
   \vspace{-4mm}
   \label{fig:method}
\end{figure*}

In this paper, we aim to align the generated images with the layout requirements in the prompts, and present a layout calibration system that requires no additional fine-tuning.
In \cref{sec:preliminaries}, we first briefly review the publicly avaliable, state-of-the-art text-to-image generator, Stable Diffusion~\cite{Rombach:Stable-Diffusion}.
In \cref{sec:check}, we introduce how to determine whether a layout correction should be initiated.
And in \cref{sec:locate}, we detail the localization of activated regions on the merged cross-attention maps.
Finally, in \cref{sec:rectify}, we present how the system rectifies the cross-attention activations according to the localized patterns and the target locations.
An overview of the pipeline is illustrated in \cref{fig:pipeline}.

\subsection{Preliminaries} \label{sec:preliminaries}

\noindent\textbf{Stable Diffusion.}
Stable Diffusion (SD)~\cite{Rombach:Stable-Diffusion} applies a hierarchical variational autoencoder (VAE)~\cite{Kingma:VAE} to operate the diffusion process~\cite{Ho:diffusion} in a low-dimensional latent space. 
Specifically, the VAE consisting of an encoder $\mathcal{E}$ and a decoder $\mathcal{D}$ is trained with a reconstruction objective.
The encoder $\mathcal{E}$ encodes the given image $\mathbf{x}$ into latent features $\mathbf{z}$, and the decoder $\mathcal{D}$ outputs the reconstructed image $\widehat{\mathbf{x}}$ from the latent, \textit{i.e.}, $\widehat{\mathbf{x}} = \mathcal{D}(\mathbf{z}) = \mathcal{D}(\mathcal{E}(\mathbf{x}))$.
To applied in a text-to-image scenario,
a pre-trained CLIP~\cite{Radford:CLIP} text encoder encodes the input textual prompt into $N$ tokens $\mathbf{y}$,
and a U-Net~\cite{Ronneberger:U-Net} consisting of convolution, self-attention, and $L$ cross-attention layers is adopted as the denoiser $\epsilon_{\theta}$.
During training, given a noised latent $\mathbf{z}^t$ and text tokens $\mathbf{y}$ at timestep $t$, the denoiser $\epsilon_{\theta}$ is optimized to remove the noise $\epsilon$ added to the latent code $\mathbf{z}$:
{\setlength\abovedisplayskip{1mm}
\setlength\belowdisplayskip{0mm}
\begin{align}
\mathcal{L}=\mathbb{E}_{\mathbf{z} \sim \mathcal{E}(\mathbf{x}), \mathbf{y}, \epsilon \sim \mathcal{N}(0,1), t}\left[\left\|\epsilon-\epsilon_\theta\left(\mathbf{z}^t, t, \mathbf{y}\right)\right\|_2^2\right].
\end{align}}%

\noindent During inference, a latent $\mathbf{z}^T$ is sampled from the standard normal distribution $\mathcal{N}(0,1)$.
At each denoising step 
$t \in [T, \cdots, 1]$, 
$\mathbf{z}^{t-1}$ is obtained by removing noise from $\mathbf{z}^{t}$ conditioned on the text tokens $\mathbf{y}$.
After the final denoising step, 
the decoder $\mathcal{D}$ maps the latent $\mathbf{z}^{0}$ to an image $\widehat{\mathbf{x}}$.

\noindent\textbf{Cross-Modal Attention.} 
The SD model leverages cross-attention layers to incorporate textual cues for the control of the image generation process.
Given the text tokens $\mathbf{y}$ and intermediate latent features $\mathbf{z}^l$, the cross-attention maps from the $l$-th layer $\mathbf{A}^l \in \mathbb{R}^{W^l \times H^l \times N}$ can be derived as

{\setlength\abovedisplayskip{-2mm}
\setlength\belowdisplayskip{2mm}
\begin{align}
\mathbf{A}^l=\text{Softmax}\left( \frac{\mathbf{Q}^l{\mathbf{K}^l}^\top}{\sqrt{d}} \right),
\end{align}}%
where $\mathbf{z}^l$ and $\mathbf{y}$ are projected to the query matrix $\mathbf{Q}$ and key matrix $\mathbf{K}$, the dimension $d$ is used to normalize the softmax values, and we omit the superscript $t$ for notational clarity and generality.
Existing studies~\cite{Chefer:Attend-and-Excite,Chen:forward-and-backward} have proposed that for the object corresponding to the $k$-th token of the prompt, higher activations on the intermediate cross-attention maps $\mathbf{A}_k^l \in \mathbb{R}^{W^l \times H^l}$ indicate the approximate position where the object will appear.
Therefore, we align the spatial location of generated objects with textual layout requirements by adjusting the activations on the cross-attention maps.

\subsection{Check}    
\label{sec:check}

A key constraint for the real-time system is to minimize the influence on the generation speed.
Therefore, \method first (1) detects the presence of object layout requirements within the text and (2) assesses any discrepancies between the generated image and the specified layout requirements.
Only if both conditions are met does the system take corrective action; otherwise, it continues with normal generation to avoid additional computational overhead.
The exact implementation of the two-step inspection is discussed below.

\noindent\faIcon[regular]{check-square} \textbf{Layout requirements exist in textual prompts.} 
Existing studies~\cite{Zheng:LayoutDiffusion,Chen:forward-and-backward} have predominantly emphasized \textbf{relative} spatial relations that are more common in written language, such as ``\textit{a dog to the left of a cat}''.
However, we argue that \textbf{superlative} spatial relations, which refer to an object shares the same relation to all other objects, have been neglected by previous research and datasets~\cite{Saharia:DrawBench}.
For example, the phrase ``\textit{a flower on the left}'' signifies that the flower is positioned to the left of all other objects, making it ideal for the leftmost target location.
In practice, it is difficult for users to directly describe their layout requirements using multiple relative expressions at once, so more direct superlative expressions actually account for a larger number.

To effectively and efficiently capture both forms of expression in a straightforward manner,
our system identifies specific positional keywords with predefined vocabulary (described in \supp).
For \textbf{relative} spatial relations, we define five spatial relations, including \textit{left}, \textit{right}, \textit{above}, \textit{below} and \textit{between}, with each relation containing a predefined vocabulary set.
And for \textbf{superlative} spatial relations, we include additional vocabulary such as ``\textit{upper-left}'' and ``\textit{lower-right}''.
The system filters out those prompts that contain words from the vocabulary set to determine the presence of layout requirements.
In practice, such a simple check implementation achieves considerable accuracy with negligible additional computational overhead.

\noindent\faIcon[regular]{check-square} \textbf{Discrepancy exists between the generated image and layout requirements.} 
To determine whether the generated image is consistent with the layout requirements, 
the target positions of all objects are necessary.
For \textbf{target layout generation}, our system provides an efficient solution by performing a dependency parsing on the prompt following with heuristic rules.
The dependency parsing can be implemented using an industrial-strength library such as spaCy~\cite{Honnibal:spaCy}.
After assigning syntactic dependency labels to tokens, \method can parse the binary ``\texttt{flower,leftmost}' from the superlative ``\textit{a flower on the left}'', and the triple ``\texttt{dog,left of,cat}'' from the relative ``\textit{a dog to the left of a cat}''.
Following pre-defined rules, the system first assigns target boxes to objects associated with superlative position terms.
Then, the remaining relative triples (and quaternions if ``\texttt{between}'' exists) can be organized as a semantic tree, with nodes as objects and edges as spatial relations.
By traversing the tree, the remaining space in the image is successively allocated.
A detailed example of assignment can be found in \supp.
For the object of the $k$-th token,
$\widehat{\mathbf{b}}_k = (\widehat{x}_k, \widehat{y}_k, \widehat{w}_k, \widehat{h}_k) \in [0, 1]^4$ denotes the assigned bounding box, where $(\widehat{x}_k, \widehat{y}_k)$ is the relative coordinates of the centre, $\widehat{w}_k$ and $\widehat{h}_k$ are the relative width and height of the box.
And the absolute boundaries $\widehat{\mathbf{b}}^l_k$ for the $l$-th layer can be computed with the concrete size of the corresponding attention map.
Note that the predicted box may not necessarily fit the size of the object and is commonly larger.
However, thanks to subsequent activation transfer, this does not affect the rectification performance.

Once the target boxes are obtained, the system prepares to assess whether each generated object is aligned with its target position.
One natural solution, using an object detector on the generated image,
requires a restart of the generation after the assessment for rectification and significantly increases the overall latency.
Therefore, 
\method places the alignment confirmation in the first denoising step (\textit{i.e.,} the $T$-th step).
Specifically, after deriving the cross-attention maps for all layers, a \textbf{layered attention merging} averages them to obtain a merged attention map:
{\setlength\abovedisplayskip{2mm}
\setlength\belowdisplayskip{2mm}
\begin{equation}
\bar{\mathbf{A}}^T=\frac{1}{L} \sum_{l=1}^L \widetilde{\mathbf{A}}^{T,l},
\end{equation}
}%
where $\widetilde{\cdot}$ means that the maps are first upsampled to a uniform resolution of $W^1 \times H^1$ before averaging.
Then, 
for the object of the $k$-th token,
\method sums over the activations within $\bar{\mathbf{A}}^T_k$ that correspond to the bounding box $\widehat{\mathbf{b}}^1_k$.
If the sum does not exceed a pre-defined threshold, the system predicts that the object will be generated in the wrong place.

\begin{figure*}[!t]    
  \centering
   \includegraphics[width=\linewidth]{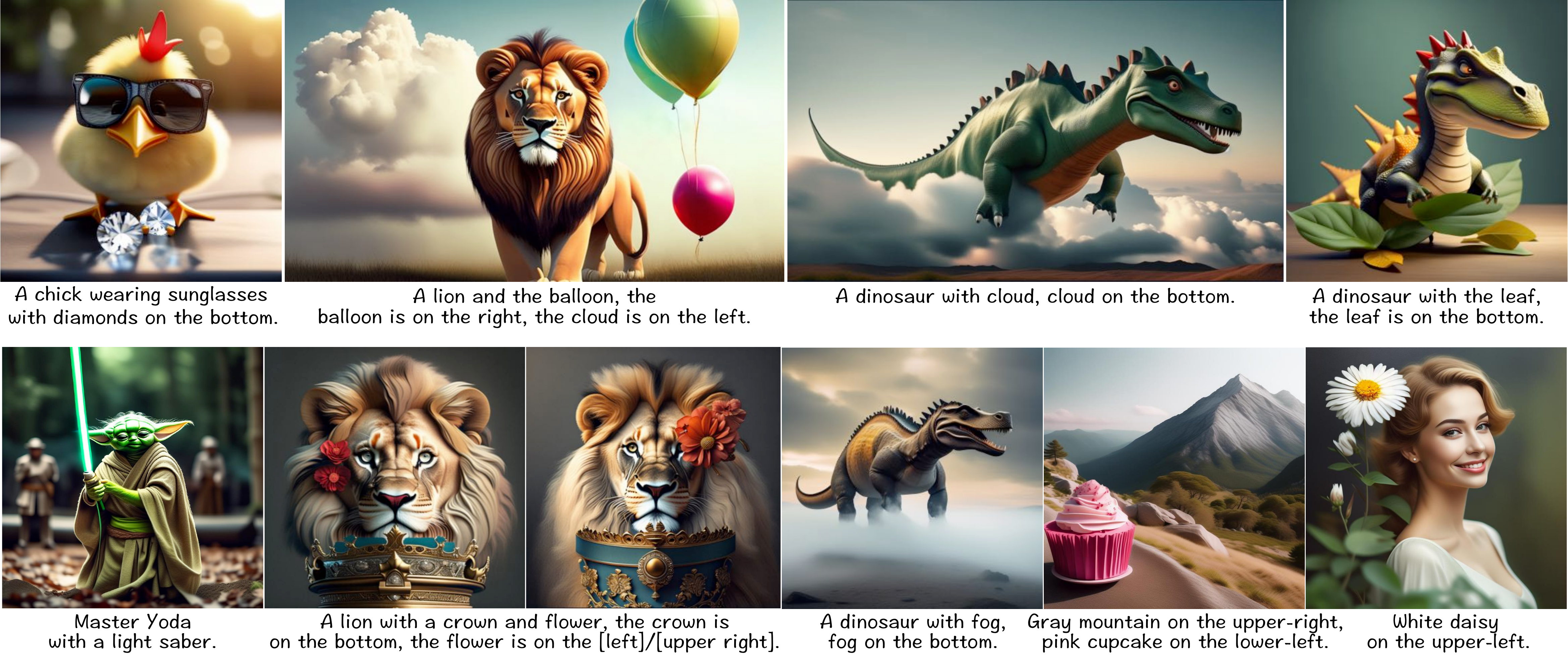}
   \vspace{-5mm}
   \caption{
   \textbf{Examples of multi-resolution image generated by \method.}
   }
   \vspace{-2mm}
   \label{fig:case2}
\end{figure*}

\subsection{Locate}
\label{sec:locate}

After confirming the initiation of the rectification, the system identifies the source activated region for each object during the early $T^{loc}$ denoising steps.

\noindent\textbf{Temporal Attention Merging.} 
For each time step $t \in [T, T-T^{loc}]$, the system simply saves the merged attention map $\bar{\mathbf{A}}^t$ without any modification.
When the $(T-T^{loc})$-th denoising step is finished, the system performs another temporal merging on all stored maps, obtaining $\bar{\mathbf{A}} \in \mathbb{R}^{W^1 \times H^1 \times N}$ that more stably indicates the source positions of generated objects:
{\setlength\abovedisplayskip{1mm}
\setlength\belowdisplayskip{1mm}
\begin{equation}
\bar{\mathbf{A}}=\frac{1}{T^{loc}} \sum_{t=T-T^{loc}}^T \bar{\mathbf{A}}^{t}.
\end{equation}
}%

\noindent\textbf{Activated Region Localization.} 
Given the temporal-merged attention map $\bar{\mathbf{A}}$, the system locates the current activated region for each object.
This is implemented by sweeping $\bar{\mathbf{A}}_k$ with a rectangular sliding window.
In practice, we keep the size of the window consistent with the target box assigned by heuristic rules.
And the activated region $\mathbf{b}^l_k$ in the $l$-th layer can be converted from the most salient window $\mathbf{b}^1_k$ found on $\bar{\mathbf{A}}_k$.

\subsection{Rectify}    
\label{sec:rectify}

After the $(T-T^{loc})$-th denoising step, the system starts to modify the generated cross-attention map for rectification.
Note that in the following statements, $\mathbf{A}$ denotes the cross-attention maps generated before applying $\text{Softmax}(\cdot)$.
Besides, the maps from the first and last cross-attention layers are not modified as we have observed that doing so improves the quality of object generation in practice.

\noindent\textbf{Activation Transfer.} 
Since the size of the localized source activated region $\mathbf{b}^l_k$ and the assigned target box $\widehat{\mathbf{b}}^l_k$ are kept the same, the activation values of the source region can be directly duplicated to the target region, while the original region is filled with minimum values.
In this way, \method easily realizes the movement of the object.
Even if the target boxes are obtained by other means (\textit{e.g.}, user-provided) rather than heuristic rules, this simple transfer remains valid after reshaping the source activated region.

\noindent\textbf{Intra-Map Activation Enhancement and Suppression.}
In practice, we have found that some objects fail to appear due to the insufficient activations in the cross-attention maps.
Also, one object may not be exactly in its target area even after the transfer.
Therefore, for the object of the $k$-th token, the system continues to modify the attention map by enhancing the activations in $\widehat{\mathbf{b}}^l_k$.
Meanwhile, to avoid the object appearing in non-target areas, the signal outside $\widehat{\mathbf{b}}^l_k$ is suppressed.
Formally, we have
\begin{equation}
\mathbf{A}^{t,l}_{k}(i,j) \leftarrow \begin{cases} 
\mathbf{A}^{t,l}_{k}(i,j) \ \cdot \alpha &\text{{if }} (i,j) \text{{ in }} \widehat{\mathbf{b}}^l_k \\
\mathbf{A}^{t,l}_{k}(i,j) \ / \ \alpha &\text{{if }} (i,j) \text{{ not in }} \widehat{\mathbf{b}}^l_k
\end{cases},
\end{equation}
where $l \in [2, L-1]$, and the hyperparameter $\alpha \in \mathbb{R}^{+}$ denotes the strength of the adjustment.

\begin{figure*}[!t]    
  \centering
   \includegraphics[width=\linewidth]{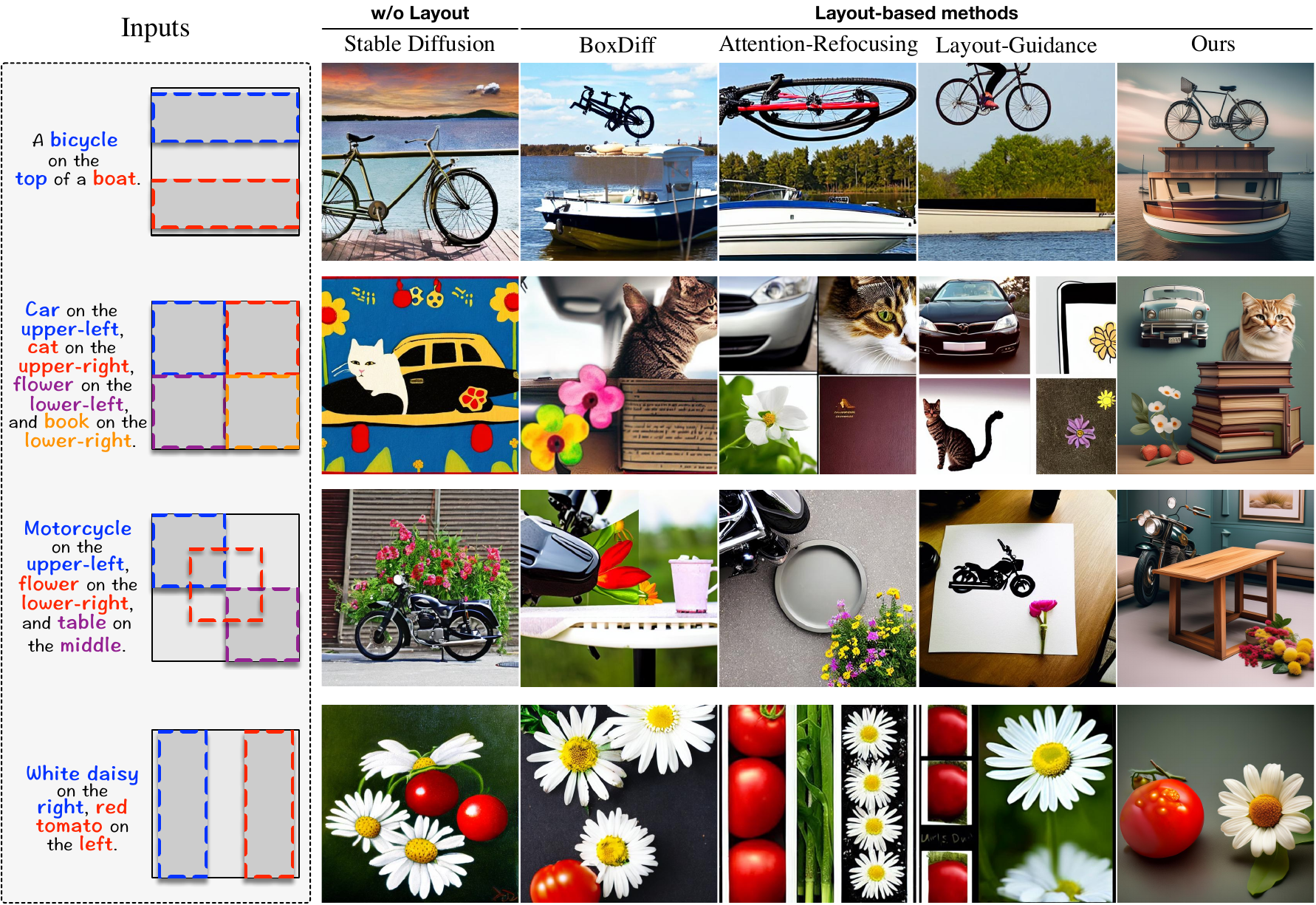}
   \caption{
   \textbf{Qualitative comparisons on DrawBench and \benchmark.}
   Textual prompts require to generate multiple objects with relative and superlative spatial relations.
   }
   \vspace{-2mm}
   \label{fig:baselines}
\end{figure*}

\noindent\textbf{Inter-Map Activation Enhancement and Suppression.}
The intra-map activation adjustment further enhances the control over the position of individual objects.
However, due to the lack of interference between attention maps, the overlap of activated areas on different maps can lead to conflict and confusion in the generation of multiple objects.
To avoid the issue, given its corresponding attention map $\mathbf{A}^{t,l}_k$ of each object, our system generates an adjustment mask $\mathbf{M}^{t,l}_k$ for other maps:
\begin{equation}
\mathbf{M}^{t,l}_k   = 1 - \text{Softmax}(\mathbf{A}^{t,l}_{k}),
\end{equation}
where the mask adjusts the attention value of other maps:
\begin{equation}
\mathbf{A}^{t,l}_{g} \leftarrow \mathbf{M}^{t,l}_k \odot \mathbf{A}^{t,l}_{g}, \ \ \text{{for }}g \in [1, N] \ \text{{and }} g \neq k.
\end{equation}
In this way, after applying $\text{Softmax}(\cdot)$, the activated regions on different maps can be staggered
to reduce conflicts.

\section{Experiments}

\noindent\textbf{Datasets.}
We utilize different datasets to evaluate the effectiveness for both relative and superlative layout requirements.
For prompts involving relative spatial relations, we use a subset of 20 prompts from the DrawBench~\cite{Saharia:DrawBench} dataset, which is a common choice of previous works~\cite{Phung:Attention-Refocusing}.
However, there is a lack of an appropriate dataset that addresses prompts concerning superlative spatial relations.
Therefore, we present a benchmark \textbf{\benchmark} consisting of 203 prompts, where each prompt contains 1 to 4 objects, and each object has superlative layout requirements.
Details are provided in~\supp.

\noindent\textbf{Baselines.}
We select Stable Diffusion~\cite{Rombach:Stable-Diffusion}, Layout-Guidance~\cite{Chen:forward-and-backward}, Attention-Refocusing~\cite{Phung:Attention-Refocusing} and BoxDiff~\cite{Xie:BoxDiff} as baselines in the main comparison.
We adopt the official implement and default hyperparameters for all baselines.

\noindent\textbf{Evaluation Metrics.}
The generation accuracy~\cite{Phung:Attention-Refocusing} is adopted as the primary evaluation metric.
Specifically, a generated image will only be considered correct if all objects are correctly generated and their spatial positions or relations, color, and other possible attributes align with the corresponding phrases in the prompt.
Following previous studies~\cite{Xie:BoxDiff}, we also report the CLIP-Score~\cite{Hessel:CLIPScore}, which measures the similarity between the input text features and the generated image features.
While this metric has been widely used to explicitly evaluate the fidelity to the text prompt, we highlight its reliability is limited, since CLIP struggles to understand spatial relationships and take them into account when scoring image-text pairs~\cite{Subramanian:ReCLIP}.

\noindent\textbf{Implementation Details.}
We adopt the DDIM scheduler~\cite{Song:DDIM} with 20 denoising steps (\textit{i.e.}, $T=20$).
And the number of localization steps $T^{loc}$ is set to 1 as default.
The ratio of classifier-free guidance is set to 5.
Adjustment strength $\alpha$ is set to 10.
Four images are randomly generated for each evaluation prompt.

\subsection{Main Results}

\begin{table}[!t]
\tablestyle{5pt}{1.0}
\setlength\tabcolsep{2pt}
\def\w{20pt} 
    \caption{%
  \textbf{Quantitative comparisons with competing methods.} 
  The generation accuracy (\%) and CLIP-Score on DrawBench~\cite{Saharia:DrawBench} and our presented \benchmark are reported.
}
\vspace{-2mm}
\scalebox{0.98}{
    \begin{tabular}{l|cc|cc}
    \multirow{2}[3]{*}{\textbf{Methods}} & \multicolumn{2}{c|}{\textbf{DrawBench}~\cite{Saharia:DrawBench}} & \multicolumn{2}{c}{\textbf{\benchmark}} \\ & Accuracy & CLIP-Score & Accuracy & CLIP-Score \\
    \shline
    Stable Diffusion~\cite{Rombach:Stable-Diffusion} & 12.50 & 0.3267 & 4.25 & 0.3012 \\
    BoxDiff~\cite{Xie:BoxDiff} & 30.00 & 0.3239 & 24.08 & \textbf{0.3032} \\
    Layout-Guidance~\cite{Chen:forward-and-backward} & 36.50 & 0.3354 & 25.50 & 0.3020 \\
    Attention-Refocusing~\cite{Phung:Attention-Refocusing} & 43.50 & 0.3339 & 50.71 & 0.3017 \\
    \rowcolor[rgb]{ .949,  .949,  .949} \method (Ours)  & \textbf{53.00} & \textbf{0.3423} & \textbf{65.16} & 0.3001 \\
    \end{tabular}%
    }
\vspace{-1mm}
  \label{tab:main_results}%
\end{table}%

\begin{figure}[!t]    
  \centering
   \includegraphics[width=\linewidth]{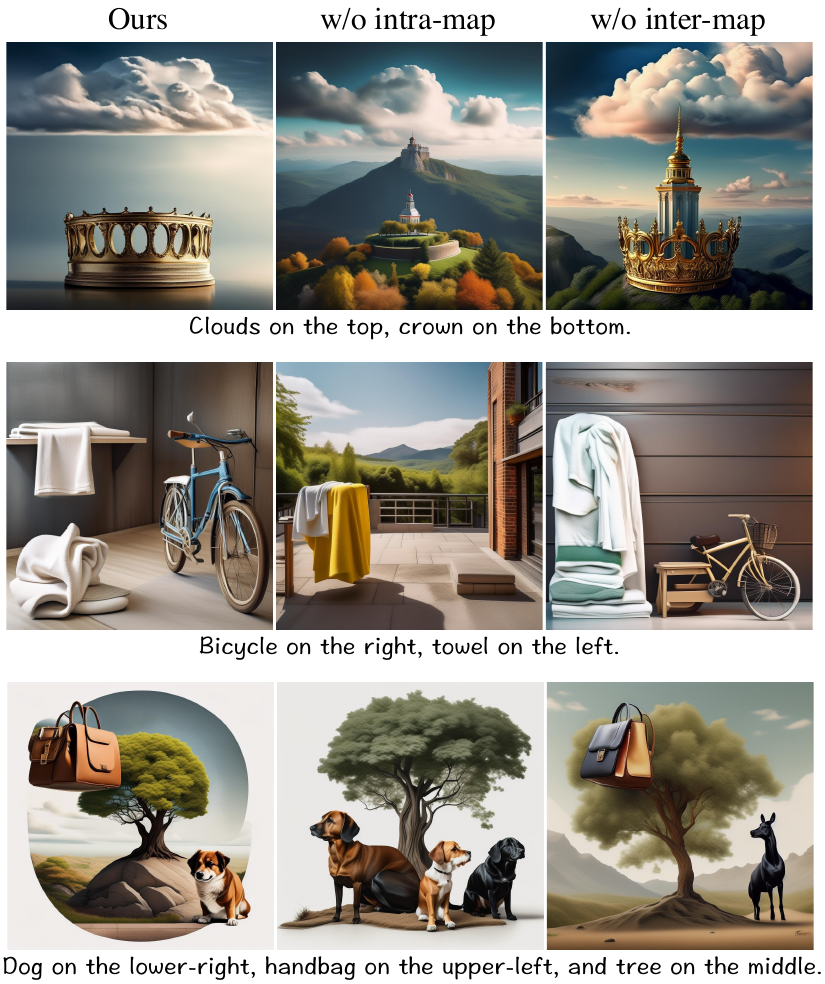}
      \vspace{-5mm}
   \caption{
   \textbf{Ablation study of intra-/inter-map activation adjustment.}
   The removal of intra-map adjustment leads to the omission of objects or positional errors, while the removal of inter-map adjustment results in fragmented or erroneous object generation.
   }
   \vspace{-3mm}
   \label{fig:ablation}
\end{figure}

\noindent\textbf{Quantitative results.}
\cref{tab:main_results} shows the quantitative comparison results between different baselines and our \method.
On the DrawBench dataset, our \method achieves the highest generation accuracy and CLIP-Score, while outperforming the baselines by a significant margin of 9.5\% in terms of accuracy.
And on the \benchmark dataset, \method not only surpasses the baselines by 14.45\% in terms of accuracy but also achieves comparable CLIP-Score.
The results signify the effectiveness of \method system in understanding both relative and superlative relationships, leading to satisfactory rectification of layout inconsistencies.

\noindent\textbf{Qualitative results.}
In \cref{fig:case2}, we present more multi-resolution images generated by \method.
\cref{fig:baselines} shows a visual comparison between the proposed \method and the competing baselines.
Without additional layout guidance, the images generated by the vanilla Stable Diffusion fail to convey the layout requirements specified by the textual prompt while also suffering from missing objects.
The three baseline models can enhance the accuracy of the generation in terms of layout.
However, they each still suffer from respective issues.
Taking the second row as an example, BoxDiff exhibits limitations in effectively controlling the layout, where the white daisies that should only appear on the right side also appear on the left and middle as well.
And the images generated by Layout-Guidance and Attention-Refocusing exhibit noticeable blockiness, tearing artifacts and object deformations, which significantly degrade the quality.
In contrast, our system maintains excellent image quality while rectifying the layout.
We attribute this to the activation localization and movement, which allows us to preserve the generative capabilities of the base model to the maximum extent, without relying on rigid constraints imposed by loss functions.

\subsection{Ablation Study}

In \cref{fig:ablation}, we visualize the generated images after removing the intra- and inter-map activation adjustments from \method.
After removing the intra-map adjustment, objects are missing (first two rows) or specified objects appear outside their target positions (the last row). 
This illustrates that the mechanism significantly contributes to controlling the placement of objects.
Meanwhile, removing the inter-map adjustment increases the likelihood of interference from activations of other maps, which can disrupt the generation of objects in their target positions, ultimately resulting in erroneous or incomplete object generation.

\begin{figure}[!t]    
  \centering
   \includegraphics[width=\linewidth]{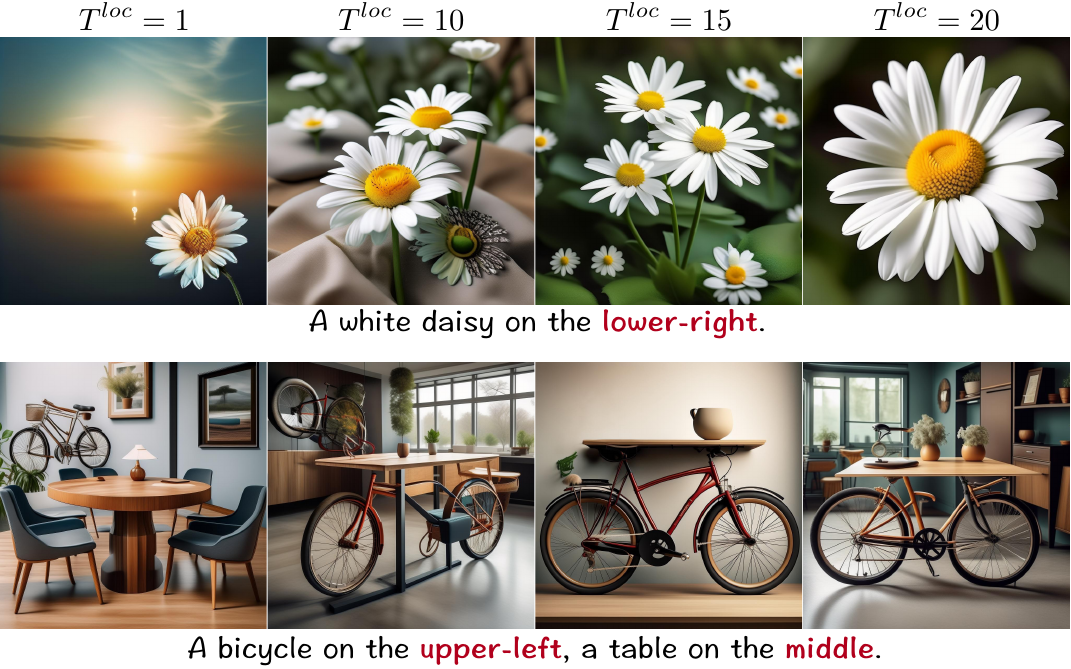}
      \vspace{-5mm}
   \caption{
   \textbf{Effect of the number of localization steps $T^{loc}$.}
   Initiating layout rectification at an earlier stage enhances the fidelity.
   }
    \vspace{-4mm}
   \label{fig:hyper_T-loc}
\end{figure}

\subsection{Further Analysis}

\noindent\textbf{Effect of the number of localization steps $T^{loc}$.}
In \cref{fig:hyper_T-loc}, we present the visual results with layout rectification initiated at different denoising steps during the generation.
It can be observed that starting the rectification from the first denoising step yields better results, ensuring that each object appears in its designated position.
The later the rectification starts, the worse the correction effect, thus compromising the fidelity of the generated images.
This observation is consistent with the conclusion from previous studies~\cite{Balaji:eDiff-I,Hertz:prompt-to-prompt-editing}, where diffusion models establish the layout in early stages and refine the appearance details in later stages.

\begin{figure}[!t]    
  \centering
   \includegraphics[width=\linewidth]{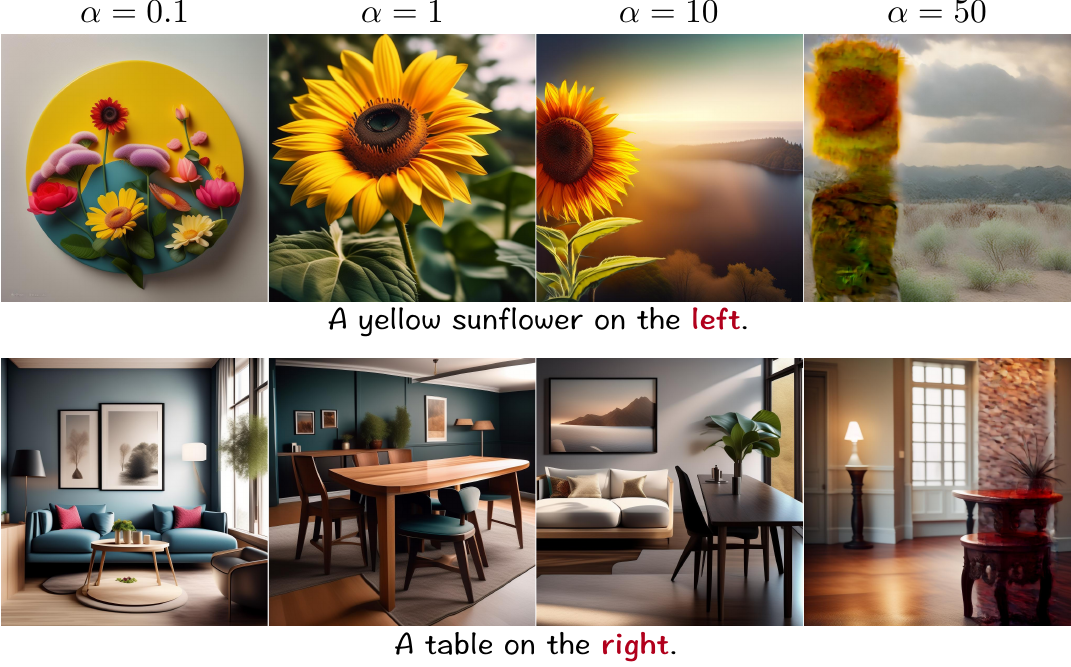}
    \vspace{-6mm}
   \caption{
   \textbf{Effect of adjustment strength $\alpha$.}
   A value of 10 yields better layout stabilization and generation quality.
   }
   \vspace{-5mm}
   \label{fig:hyper_alpha}
\end{figure}

\noindent\textbf{Effect of adjustment strength $\alpha$.} 
We scale $\alpha$ from 0.1 to 50 and illustrate some generated cases in \cref{fig:hyper_alpha}.
Setting $\alpha$ to 0.1 essentially reverses the enhancement and suppression, resulting in objects appearing in non-designated positions.
And setting $\alpha$ to 1 essentially removes the intra-map attention adjustment, leading to less effective layout rectification.
Further increasing the $\alpha$ to 10 yields facilitates rectification and provides better control over the layout.
However, excessively large values of $\alpha$ (\textit{e.g.}, setting it to 50) can degrade the quality of the generated images while imposing stricter constraints on the object positions.

\section{Related Work}

\noindent\textbf{Text-to-Image Generation.}
Earlier works studied text-to-image generation in the context of generative adversarial networks (GANs)~\cite{Reed:GAN-T2I,Xu:AttnGAN,Tao:DF-GAN,Zhu:DM-GAN}.
Despite their dominance, the adversarial training nature brings the issues including training instability and less diversity in generation~\cite{Dhariwal:diffusion-beats-GAN}.
Text-conditional auto-regressive models~\cite{Ramesh:DALL-E,Gafni:Make-A-Scene,Ding:CogView,Yu:Parti} demonstrated more impressive results while requiring time-consuming iterative processes to achieve high-quality image sampling.
Natural fitting to inductive biases of image data, the emerging diffusion models~\cite{Saharia:Imagen,Ramesh:DALL-E2,Nichol:GLIDE,Rombach:Stable-Diffusion} have recently demonstrated impressive generation results based on open-vocabulary text descriptions.
To reduce training overhead and speed up inference, latent diffusion model~\cite{Rombach:Stable-Diffusion} trims off pixel-level redundancy by applying an autoencoder to project images into latent space and generating latent-level feature maps with the diffusion process.
And to align with the provided textual input, Stable Diffusion~\cite{Rombach:Stable-Diffusion} further employs cross-attention mechanism to inject textual condition into the diffusion generation process.

\noindent\textbf{Layout Control in Diffusion Models.}
Existing progress fails to fully understand the spatial relations of objects in the free-form textual descriptions and reflect them in the synthesized image, especially for complex scenes.
Therefore, jointly conditioning on text and layout has been studied, where layout control signals can be bounding boxes~\cite{Qu:LayoutLLM-T2I,Xie:BoxDiff}, segmentation maps~\cite{Xue:FLIS,Couairon:ZestGuide,Avrahami:SpaText}, and key points~\cite{Zhang:continuous-layout-editing}.
Several methods extend the Stable Diffusion model by incorporating layout tokens into attention layers~\cite{Li:GLIGEN,Yang:ReCo,Zheng:LayoutDiffusion} or training layout-aware adapters~\cite{Qu:LayoutLLM-T2I}.
However, requiring additional training on massive layout-image pairs, these approaches lack flexibility in the base model and may degrade the quality of the generated images.
Therefore, recent efforts~\cite{Chen:forward-and-backward,Phung:Attention-Refocusing,Xie:BoxDiff} design loss conditioned on layout constraints to update the noised latent together with denoising.
Layout-Guidance~\cite{Chen:forward-and-backward} computes the loss by applying the energy function on the cross-attention map,
Attention-Refocusing~\cite{Phung:Attention-Refocusing} constrains both cross-attention and self-attention to ``refocus'' on the correct regions, and
BoxDiff~\cite{Xie:BoxDiff} designs inner-box, outer-box, and corner spatial constraints.
However, they introduce extra computational cost for gradient update, which affects the speed of generation.
In contrast, our system directly modifies the activations to conform to the target for rectification, minimizing the computation overhead.

\noindent\textbf{Layout Generation.}  \label{sec:related-layout-generation}
Previous layout-to-image studies~\cite{Zheng:LayoutDiffusion,Chen:forward-and-backward} have largely neglected the discussion on layout generation and heavily relied on users to directly provide accurate layout boxes for objects.
However, this necessitates assessing the legality of user input and increases the learning and interaction difficulty for users.
Moreover, we have observed a substantial decline in the quality of generated images when the provided boxes are insufficiently accurate.
Latest efforts~\cite{Phung:Attention-Refocusing,Qu:LayoutLLM-T2I,Lian:LLM-grounded-Diffusion} have turned to large language models like GPT-4~\cite{OpenAI:GPT-4} by creating appropriate prompting templates to generate layouts, while each API request adds response time and incurs additional costs.
In this paper, our system provides a light-weight solution based on dependency parsing following with heuristic rules.
\section{Conclusion}
In this paper, we propose a training-free layout calibration system \method for text-to-image generators, 
which aligns the synthesized images with layout instructions in a post-remedy manner.
Following a ``check-locate-rectify'' pipeline, \method first decides whether to perform the layout rectification by checking the input prompt and the intermediate cross-attention maps.
During the rectification, the system identifies and relocates the activations of mispositioned objects, where the target positions are generated by analysing the prompt with dependency parsing and heuristic rules.
To comprehensively evaluate the effectiveness of \method, we present a benchmark called \benchmark, which covers both simple and complex layouts described in terms of superlative relations.
Through extensive qualitative and quantitative experiments, we demonstrate our superiority in improving generation fidelity and quality.


{
    \small
    \bibliographystyle{ieeenat_fullname}
    \bibliography{main}
}
\clearpage
\setcounter{page}{1}
\maketitlesupplementary

\appendix

\renewcommand{\thesection}{\Alph{section}}

\begin{figure*}[hb]    
  \centering
   \includegraphics[width=\linewidth]{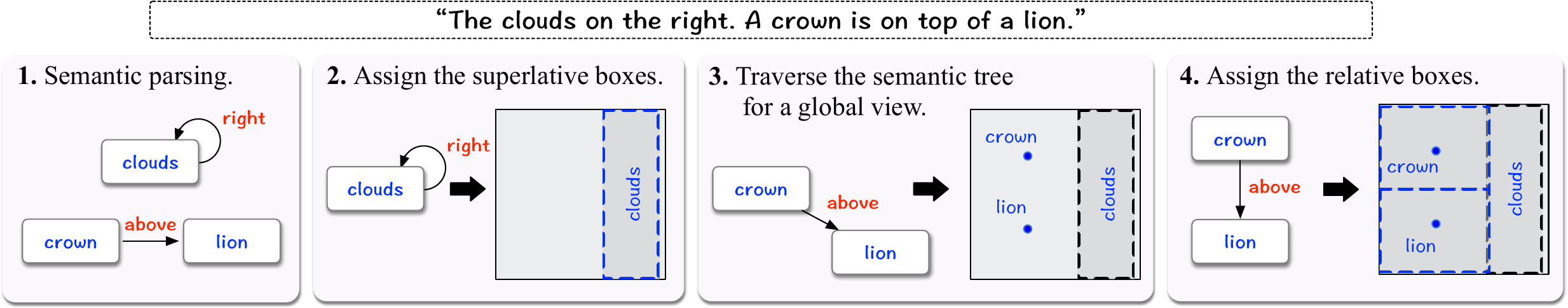}
   \caption{
   \textbf{Example of target layout generation.}
   }
   \label{fig:layout_generation}
\end{figure*}

\section{Relation Vocabulary for Checking}  \label{sec:vocabulary}

Our \method determines the existence of layout requirements by checking whether any words from our predefined relation vocabulary are present in the prompt.
According to the semantic similarity,
the vocabulary contains six categories:

\begin{itemize}
    \item \textbf{left}: ``\textit{left}'', ``\textit{west}''
    \item \textbf{right}: ``\textit{right}'', ``\textit{east}''
    \item \textbf{above}: ``\textit{above}'', ``\textit{over}'', ``\textit{on}'', ``\textit{top}'', ``\textit{north}''
    \item \textbf{below}: ``\textit{below}'', ``\textit{beneath}'', ``\textit{underneath}'', ``\textit{under}'', ``\textit{bottom}'', ``\textit{south}''
    \item \textbf{between}: ``\textit{between}'', ``\textit{among}'', ``\textit{middle}''  
    \item \textbf{additional superlative}: ``\textit{upper-left}'', ``\textit{upper-right}'', ``\textit{lower-left}'', ``\textit{lower-right}''
\end{itemize}

\noindent Note that 
(1) The ``additional superlative'' category serves as a supplement for words that have not been covered.
In the given context, words such as ``\textit{left}'' and ``\textit{above}'' can also 
represent the superlative relations.
(2) This vocabulary can easily be extended according to the needs of the dataset.

\section{Superlative Predefined Positions}  \label{sec:superlative_boxes}

For each object associated with a superlative relation, the relative bounding box $\widehat{\mathbf{b}} = (\widehat{x}, \widehat{y}, \widehat{w}, \widehat{h})$ is assigned as follows:

\begin{itemize}
    \item \textbf{left}: (0.20, 0.50, 0.33, 1.00)
    \item \textbf{right}: (0.80, 0.50, 0.33, 1.00)
    \item \textbf{above}: (0.50, 0.20, 1.00, 0.33)
    \item \textbf{below}: (0.50, 0.80, 1.00, 0.33)
    \item \textbf{middle}: (0.50, 0.50, 0.50, 0.50)
    \item \textbf{upper-left}: (0.25, 0.25, 0.50, 0.50)
    \item \textbf{upper-right}: (0.75, 0.25, 0.50, 0.50)
    \item \textbf{lower-left}: (0.25, 0.75, 0.50, 0.50)
    \item \textbf{lower-right}: (0.75, 0.75, 0.50, 0.50)
\end{itemize}

\section{An Example of Target Layout Generation}  \label{sec:heuristic-rules}

To facilitate understanding of how \method parses the prompt and generates the target bounding box for each object with a set of heuristic rules, 
we show an example in \cref{fig:layout_generation} to illustrate it more clearly.
Specifically, the process can be roughly divided into four steps:

\noindent \textbf{1. Semantic parsing.} \method parses the superlative tuples and relative triplets from the prompt. And the relative triples can be organized as a semantic tree, with nodes as objects and edges as spatial relations. 

\noindent \textbf{2. Assign the superlative boxes.} Given each superlative tuple, \method assigns a predefined target box to the object according to its superlative position term.

\noindent \textbf{3. Traverse the semantic tree for a global view.} By traversing the tree, \method organizes the global layout of the remaining objects.

\noindent \textbf{4. Assign the relative boxes.} \method allocates the remaining space to the objects associated with superlative relations.

\begin{table}[!t]
\tablestyle{5pt}{1.0}
\setlength\tabcolsep{2pt}
\def\w{20pt} 
    \caption{%
  \textbf{Detailed quantitative results on \benchmark.} 
  The generation accuracy (\%) is reported.
}
\vspace{-2mm}
\scalebox{1}{
    \begin{tabular}{l|c|c|c|c}
    Methods & 1 object & 2 objects & 3 objects & 4 objects \\
    \shline
    Stable Diffusion~\cite{Rombach:Stable-Diffusion} & 15.56 & 5.21  & 0.00  & 0.00 \\
    BoxDiff~\cite{Xie:BoxDiff} & 41.11 & 18.23 & 19.64 & 13.33 \\
    Layout-Guidance~\cite{Chen:forward-and-backward} & 82.22 & 5.73  & 3.57  & 20.00 \\
    Attention-Refocusing~\cite{Phung:Attention-Refocusing} & 65.56 & 41.67 & 57.14 & 53.33 \\
    \rowcolor[rgb]{ .949,  .949,  .949} \method (Ours)  & \textbf{82.22} & \textbf{53.64} & \textbf{76.79} & \textbf{66.67} \\
    \end{tabular}%
    }
\vspace{-2mm}
  \label{tab:detailed_results}
\end{table}%

\begin{figure*}[!t]    
  \centering
   \includegraphics[width=\linewidth]{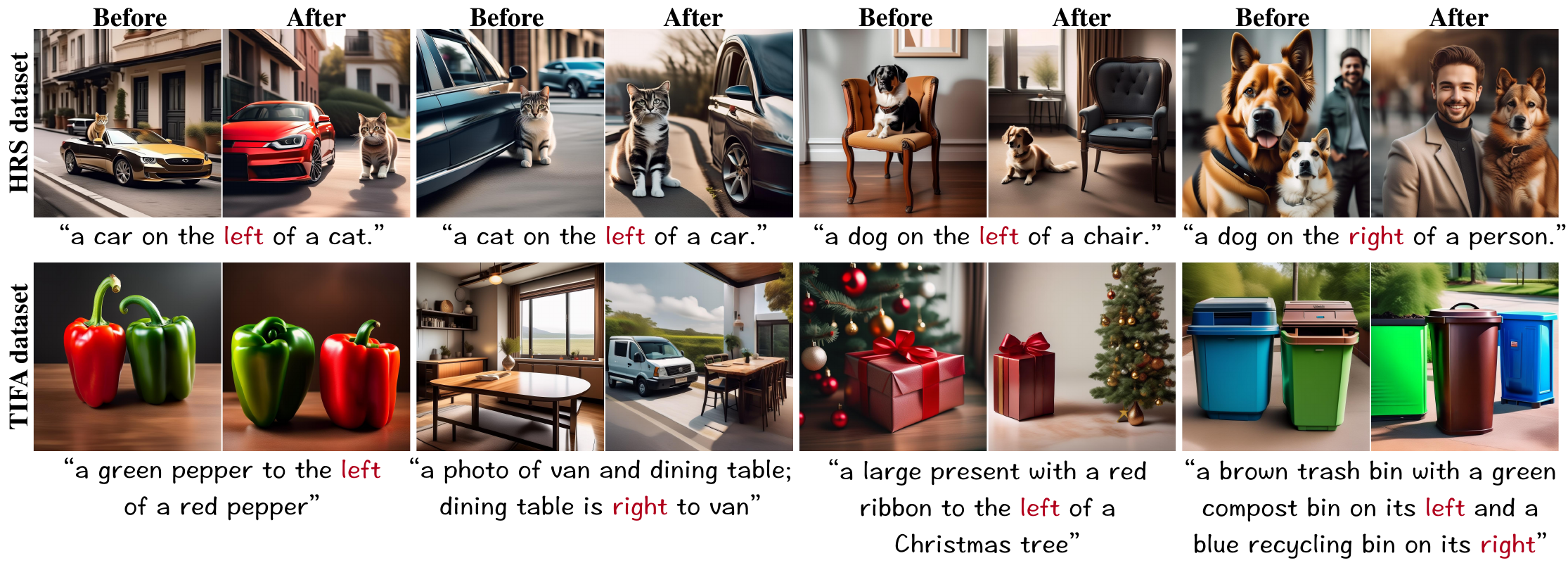}
   \caption{
   \textbf{Qualitative results on HRS~\cite{Bakr:HRS-Bench} and TIFA~\cite{Hu:TIFA} benchmarks.}
   }
   \label{fig:HRS_TIFA}
\end{figure*}

\section{Benchmark Details}  \label{sec:benchmark-details}

\textbf{Overview.}
Our proposed \benchmark focuses on superlative relations.
Specifically, to sample an evaluation prompt, we first determine the number of objects in the prompt.
Each prompt contains a minimum of one object and a maximum of four objects.
Then, we sample the superlative relation for each object that has not yet been determined, where the predefined superlative relation set is the same as shown in \cref{sec:superlative_boxes}.
Finally, we sample the objects present in the current prompt from a predefined set of objects.
To better evaluate the impact of layout requirements on image generation, a sampled object set can be shared between prompts with different superlative relations.
As a result, \benchmark contains 203 different prompts.
The number of prompts containing 1/2/3/4 objects is 36/96/56/15.
And the number of occurrences of each superlative relation is 55/55/49/49/56/48/48/48/48.
The benchmark will be publicly available.

\noindent \textbf{Object set.}
The predefined object set consists of 28 different items as follows:

\begin{itemize}
    \item \textbf{single-word}: ``\textit{backpack}'', ``\textit{flower}'', ``\textit{crown}'', ``\textit{towel}'', ``\textit{scarf}'', ``\textit{beach}'', ``\textit{clouds}'', ``\textit{tree}'', ``\textit{table}'', ``\textit{book}'', ``\textit{handbag}'', ``\textit{bus}'', ``\textit{bicycle}'', ``\textit{car}'', ``\textit{motorcycle}'', ``\textit{cat}'', ``\textit{dog}'', ``\textit{horse}''
    \item \textbf{phrase}: ``\textit{chocolate cookie}'', ``\textit{strawberry cake}'', ``\textit{vanilla ice cream cone}''
    \item \textbf{with color}: ``\textit{yellow sunflower}'', ``\textit{gray mountain}'', ``\textit{white daisy}'', ``\textit{pink cupcake}'', ``\textit{red tomato}'', ``\textit{golden saxophone}'', ``\textit{green broccoli}''
\end{itemize}

\section{Detailed Accuracies on \benchmark}

In \cref{tab:detailed_results}, we report the accuracies when the number of objects in the prompt is different.
It can be observed that our \method outperforms the baselines in all cases.
Furthermore, despite the simplicity of the case with a single object, the accuracies do not show a clear downward trend as the number of objects increases.
The difficulty of accurately representing the layout is also influenced by the specific layout requirements of the objects and their context.

\section{Additional Results}

\subsection{Latency Comparison for Layout Generation}

Since our \method system presents a new solution for generating the target layout, we provide a brief discussion of the observed increase in latency here.
Existing layout-to-image works~\cite{Phung:Attention-Refocusing,Qu:LayoutLLM-T2I} commonly rely on GPT-4~\cite{OpenAI:GPT-4}, however, each invocation of the API requires a response time of
$\sim$3 seconds.
In contrast, thanks to the industrial-strength library, our proposed solution requires an average of only 0.006 seconds for each prompt and does not require a GPU.
This significantly improves the user experience for real-time text-to-image generators.

\begin{figure}[!t]    
  \centering
   \includegraphics[width=\linewidth]{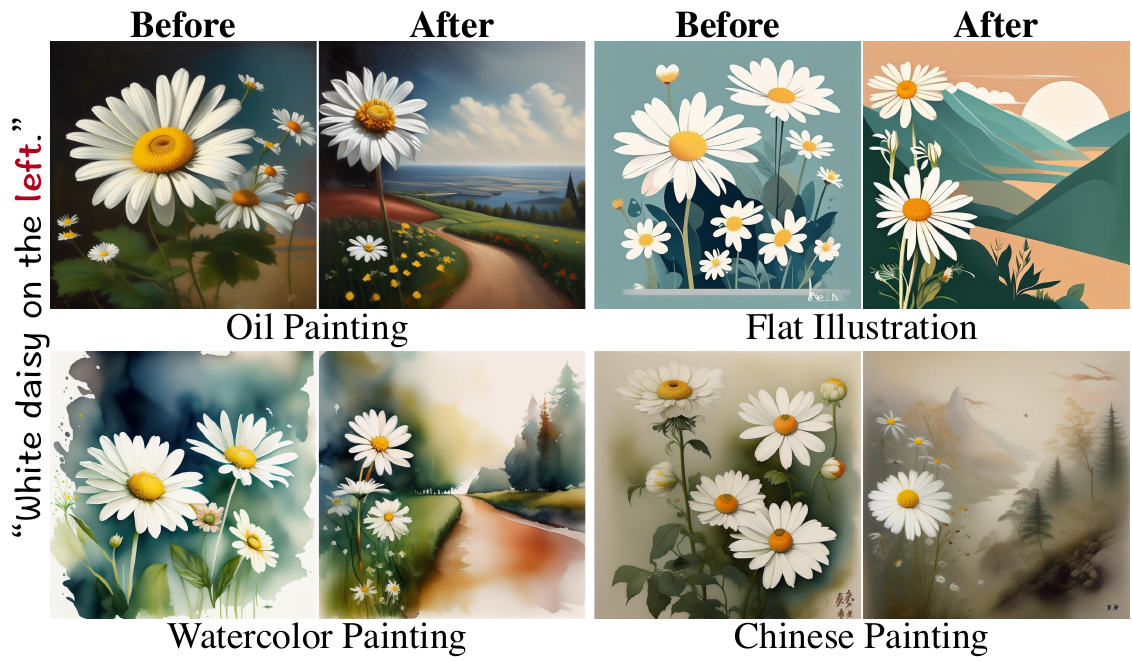}
   \caption{
   \textbf{Layout calibration results of images in different styles.}
   }
   \label{fig:diff_styles}
\end{figure}

\subsection{Generalization Across Diverse Styles}

In practical scenarios, users often request the text-to-image generators to produce images in specific styles.
In \cref{fig:diff_styles}, we show that the stylistic demands for generated images do not hinder the rectification of the layout by \method.

\begin{figure}[!b]    
  \centering
  \vspace{-5mm}
   \includegraphics[width=0.9\linewidth]{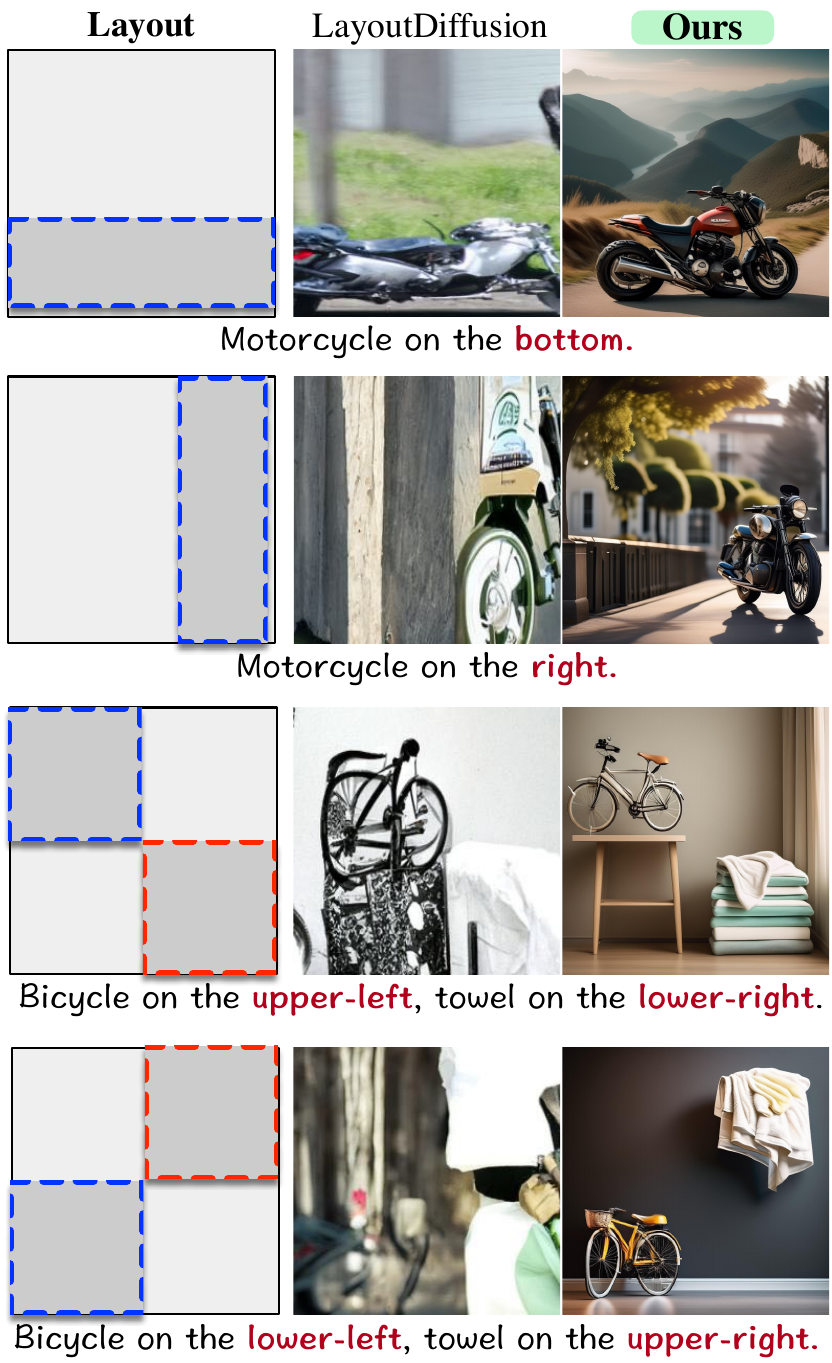}
   \caption{
   \textbf{Qualitative comparisons with LayoutDiffusion~\cite{Zheng:LayoutDiffusion}.}
   The generation quality of LayoutDiffusion is far worse than \method.
   }
   \label{fig:vs-LayoutDiffusion}
\end{figure}

\subsection{Qualitative Results on Other Benchmarks}

We additionally present the qualitative results obtained on two latest benchmarks, HRS~\cite{Bakr:HRS-Bench} and TIFA~\cite{Hu:TIFA}, in \cref{fig:HRS_TIFA}.
These two benchmarks, similar to DrawBench, excessively focus on relative spatial relations.
Due to the cost of comprehensive manual evaluation, we take quantitative evaluation on these benchmarks as future work.

\subsection{Comparison with Training-Based Method}

LayoutDiffusion~\cite{Zheng:LayoutDiffusion} is a representative approach in training auxiliary modules to embed the layout information into intermediate features for controlling.
However, it is constrained to fixed categories, thereby rendering it unsuitable for various datasets including Drawbench.
To compare our \method with LayoutDiffusion, we select prompts that only includes valid objects for LayoutDiffusion from our \benchmark.
As observed in \cref{fig:vs-LayoutDiffusion}, 
the limitation of layout significantly reduces the generation quality of LayoutDiffusion, resulting in its performance being far inferior to \method.

\subsection{Failure Case Analysis}

\begin{figure}[!t]    
  \centering
   \includegraphics[width=0.8\linewidth]{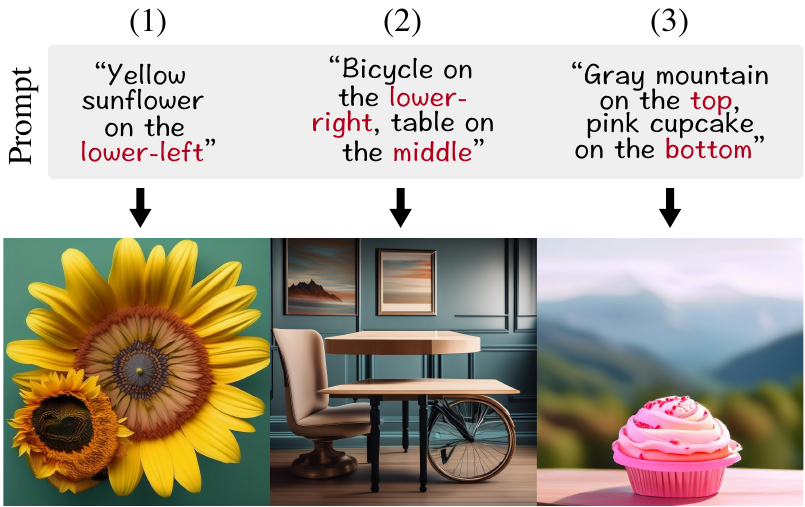}
   \caption{
   \textbf{Typical failure cases identified by human evaluators.}
   }
   \label{fig:failure_cases}
\end{figure}

In \cref{fig:failure_cases}, we present typical cases of what the human evaluators perceive as errors.
The first case is the repeated generation of objects with some in the wrong position. 
The second case is that multiple objects interact with each other during generation, resulting in incomplete generation.
The third case includes missing or unclear objects.
These errors are mostly due to the fact that a single adjustment strength parameter $\alpha$ may not be optimal for all generation.
This results in insufficient activation enhancement or suppression on the attention map, leading to inaccuracies in generating all objects accurately or preventing the repeated generation.

\subsection{Additional Qualitative Comparison Results}

\begin{figure*}[!t]    
  \centering
   \includegraphics[width=0.93\linewidth]{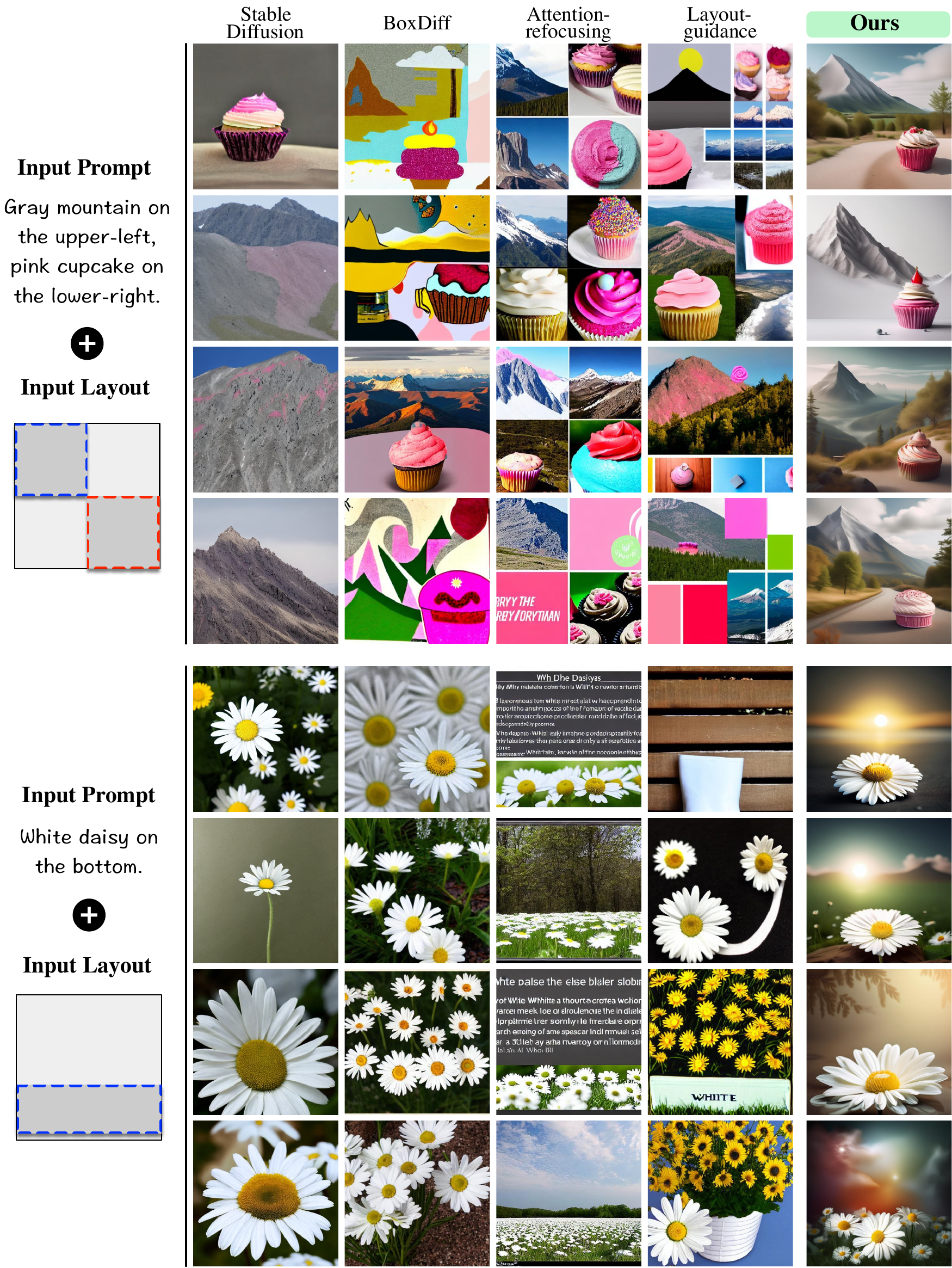}
   \caption{
   \textbf{Additional qualitative comparisons.}
   }
   \label{fig:addtional-4-1}
\end{figure*}

\begin{figure*}[!t]    
  \centering
   \includegraphics[width=0.93\linewidth]{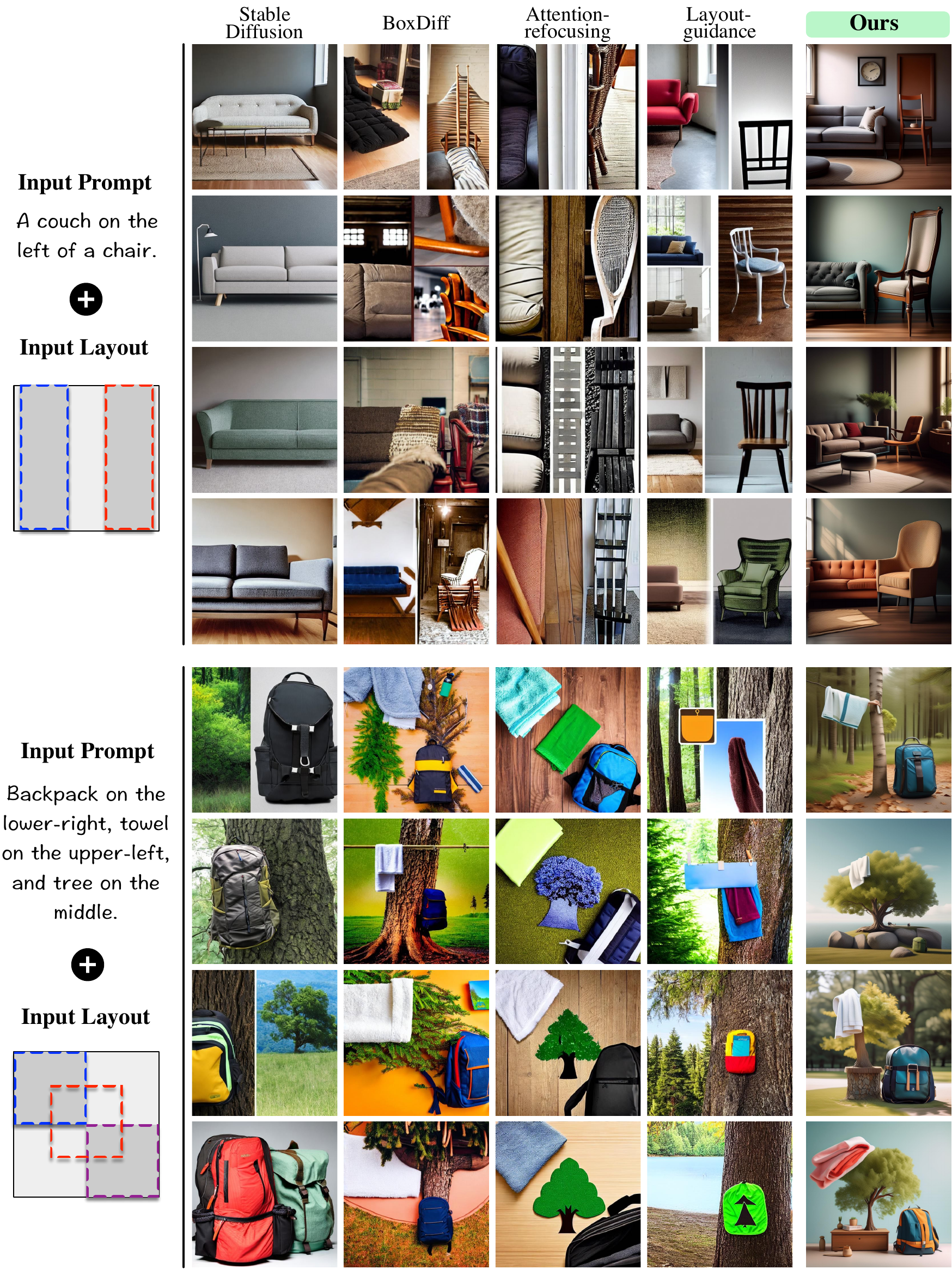}
   \caption{
   \textbf{Additional qualitative comparisons.}
   }
   \label{fig:addtional-4-2}
\end{figure*}

To show the effectiveness of \method, 
we illustrate additional qualitative results in \cref{fig:addtional-4-1,fig:addtional-4-2}.

\end{document}